\documentclass[runningheads,a4paper]{llncs}

\usepackage{amssymb}
\usepackage{graphicx}

\usepackage{url}
\urldef{\mailsa}\path|kkarpou@cs.ntua.gr|
\urldef{\mailsb}\path|gtsatiris@image.ntua.gr|
    
\newcommand{\keywords}[1]{\par\addvspace\baselineskip
\noindent\keywordname\enspace\ignorespaces#1}

\begin{document}

\mainmatter  

\title{AI in (and for) Games}

\titlerunning{AI in (and for) Games}

%
%
\author{Kostas Karpouzis\inst{1} \and George Tsatiris\inst{2}}
\authorrunning{K. Karpouzis, G, Tsatiris}

\institute{Panteion University of Social and Political Sciences, Athens, Greece,\\
\email{kkarpou@cs.ntua.gr},
\and
Artificial Intelligence and Learning Systems Laboratory,\\National Technical University of Athens, Greece}

%
%

\toctitle{Lecture Notes in Computer Science}
\tocauthor{Authors' Instructions}
\maketitle

\begin{abstract}
This chapter outlines the relation between artificial intelligence (AI) / machine learning (ML) algorithms and digital games. This relation is two-fold: on one hand, AI/ML researchers can generate large, in-the-wild datasets of human affective activity, player behaviour (i.e. actions within the game world), commercial behaviour, interaction with graphical user interface elements or messaging with other players, while games can utilise intelligent algorithms to automate testing of game levels, generate content, develop intelligent and responsive non-player characters (NPCs) or predict and respond player behaviour across a wide variety of player cultures. In this work, we discuss some of the most common and widely accepted uses of AI/ML in games and how intelligent systems can benefit from those, elaborating on estimating player experience based on expressivity and performance, and on generating proper and interesting content for a language learning game.
\end{abstract}

\section{Introduction}
Digital games have enjoyed a huge wave of popularity among the research community in the past years. An important reason for this is the fact that games combine the characteristics and requirements of performance/narrative media \cite{carstensdottir2019player} with high-maintenance software and hardware requirements for storage, CPU performance, network communication \cite{sabet2020towards} and security \cite{min2019security}. Especially in the field of computer hardware, digital games have been a driving force for the industry to produce newer, more effective and less power-demanding hardware for PCs and game consoles \cite{JPR}, pushing their capabilities further and further.

Another important fact that makes digital games extremely popular as a research medium has to do with the relative ease to find participants for games research studies: Pew Research \cite{Pew} mentions that ``43\% of U.S. adults say they often or sometimes play video games on a computer, TV, game console or portable device'', with puzzle and strategy games constituting the most popular genres among those who often or sometimes play video games. As a result, researchers working with games, either in the core of their work or as a means to attract users and record their behaviour or preferences, can quickly put together large corpora of data (e.g. \cite{karpouzis2015platformer} or \cite{shaker2011game} for a database which captures player expressivity associated with player behaviour or \cite{gourgari2013thetis} for a 3D dataset describing tennis-related actions in video and 3D skeleton form).

Among the research areas that embraced digital games as a platform of choice, perhaps the most celebrated one has been the combination of Artificial Intelligence (AI) and Machine Learning (ML), mostly because of the popularity of AI/ML algorithms which competed against and eventually beat human world champions in Chess \cite{campbell2002deep} and more complex board games such as Go (\cite{silver2016mastering}, \cite{silver2018general}). Games are a fitting medium to train and test AI/ML algorithms because of the relatively small search space in which to look for and identify the best possible turn and, mostly, for the completeness and robustness of the definition of the game world in terms of variables, rules and relations. Conversely, game design and development has been putting to use AI/ML algorithms to create content automatically or in a user-guided manner, to estimate and adapt player experience (\cite{guckelsberger2017predicting}, \cite{wiemeyer2016player}) or predict player behaviour \cite{charles2020behavlet}. In this chapter, we will start with identifying possible sources of data to be used to train and test AI/ML algorithms; in the following, we will discuss the areas of conversation between intelligent algorithms and digital games, providing examples of identification of player behaviour and prediction of player experience, and elaborate on game content generation for serious games in education.

\section{Game content and databases}
\begin{figure}[h]
    \centering
    \includegraphics[width=0.9\textwidth]{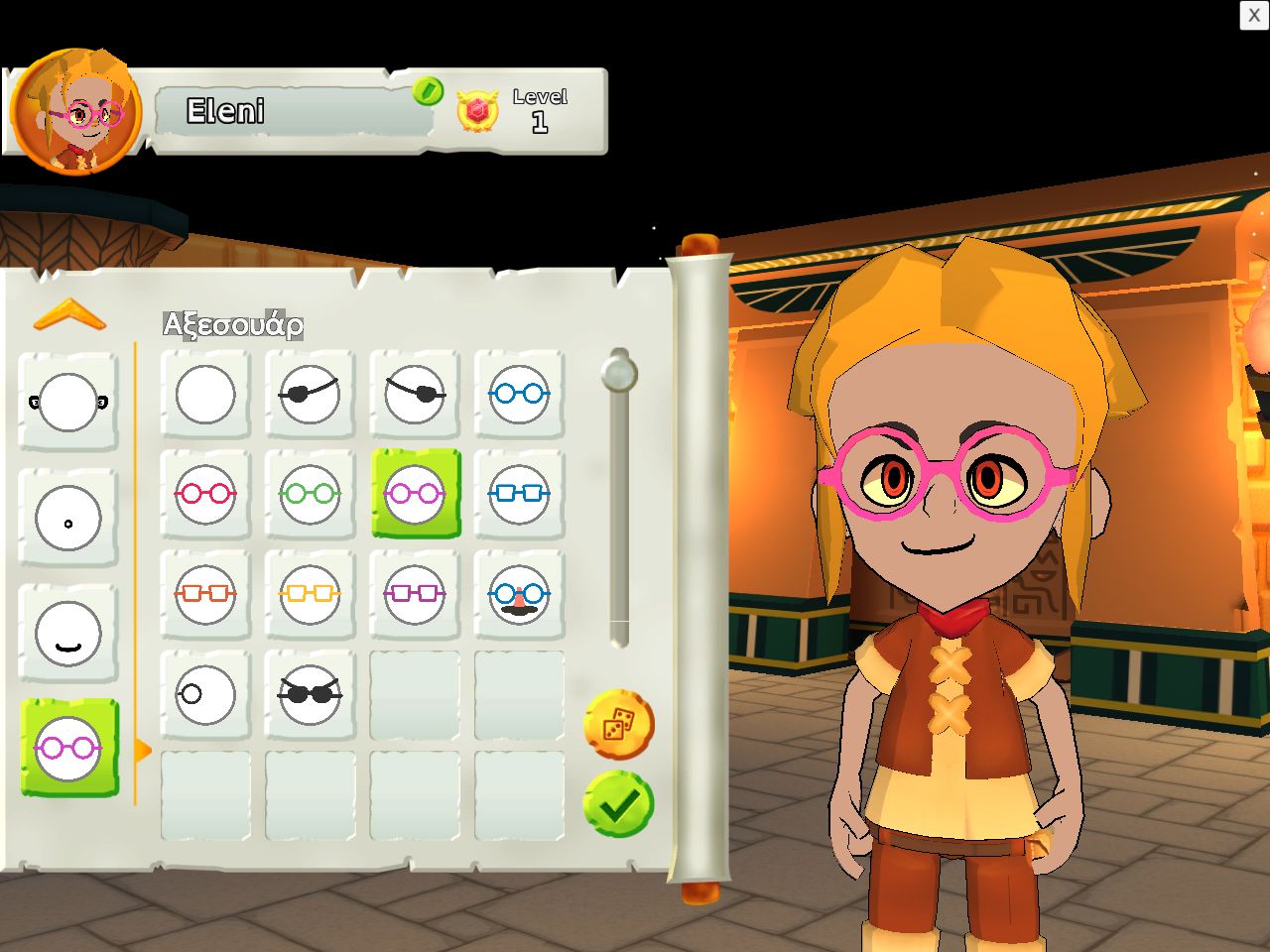}
    \caption{Example of User Generated Content, where players choose the appearance of their avatar}
    \label{personalisation}
\end{figure}

Perhaps the most important component for the interplay between AI/ML, along with the actual algorithms and their context or use case, is the selection and role of data or content to be used or generated. In game design and development terminology, \emph{content} refers to a wide variety of concepts, data and types of media within a game world. An interesting distinction is that besides content being included in the game as part of its design or interactive functionality, it can be generated by the game during game play \cite{amato2017procedural}, usually as a response of the game world to the choices and actions performed by the player, or it can be created by the players themselves \cite{lastowka2007user}. The latter case is usually referred to as ``user'' (UGC) or ``player-generated content'' and covers personalisation of the appearance of the player character (PC) in the game world (Figure \ref{personalisation}), choices that relate to game actions \cite{tsihrintzis2019machine}, commercial behaviour (e.g. buying digital goods or aesthetic elements using real money or in-game digital currency) or interaction with other players, usually using text chat, voice communication or even sign language \cite{caridakis2014non} capabilities.

\begin{figure}[h]
    \centering
    \includegraphics[width=0.85\textwidth]{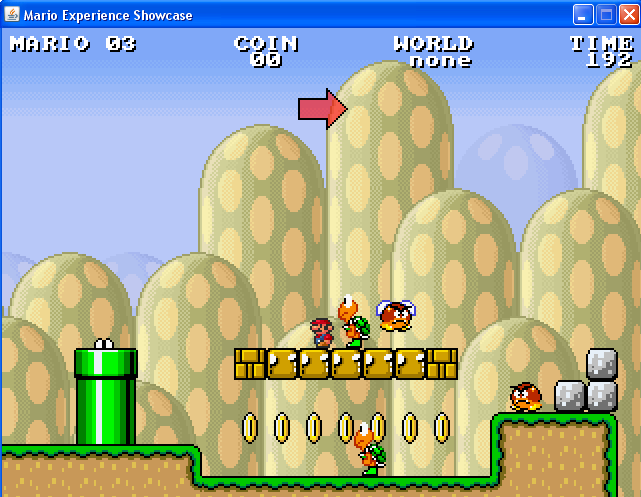}
    \caption{Generated game level in IMB, used during the recording of the PED dataset}
    \label{mario}
\end{figure}

An interesting trait of player-generated content is that, as research shows \cite{cairns2013but}, it is less controlled by social filters and inhibition; this effectively means that players who participate and function within the safe sandbox of a digital game world express themselves more vividly \cite{durning2016gaming}, using a wider range of spontaneous emotions \cite{kotsia2016multimodal} and microexpressions \cite{hemenover2018video} than usual affect- and emotion-related interactions \cite{karpouzis2016emotion}, thus offering richer input for AI/ML algorithms and catering for their deployment ``in-the-wild'' (\cite{perron2005cognitive}, \cite{cowie2011issues}). An example of spontaneous user-generated content was presented in the Platformer Experience Dataset (or PED\footnote{The dataset can be downloaded from https://ped.institutedigitalgames.com/}) \cite{karpouzis2015platformer}. This is a multimodal dataset which contains videos of 58 participants playing IMB \cite{togelius20102009}, an open-source clone of the popular ``Super Mario Bros'' platformer game (Figure \ref{mario}), recordings of the game screen synchronised with the videos, logs of player actions with timestamps and a self-reported assessment of fun, interest and player experience in two forms, ratings and ranks (see \cite{yannakakis2017ordinal} for an interesting discussion on \emph{ranking} different choices or preferences, instead of \emph{rating} each of them).

\begin{figure}[h]
    \centering
    \includegraphics[width=0.8\textwidth]{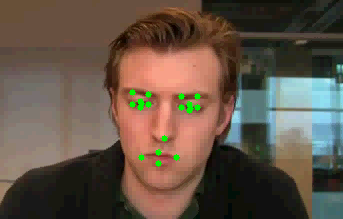}
    \caption{Video frame showing player expressivity and detected facial features}
    \label{expressivity}
\end{figure}

The approach of recording player expressivity along with player behaviour (actions in the game world) and player experience allows for a number of possible uses of AI/ML algorithms, either on recognition/classification or on generation of game content. For example, Asteriadis et al. combined head and body expressivity from players (Figure \ref{expressivity}) with in-game actions to cluster them in different groups \cite{asteriadis2012towards}, taking also game performance and demographics into account \cite{asteriadis2012does}. An interesting observation of this work was that players often used microexpressions or microgestures in conjunction with their actions and the relevant movement of their character; for instance, they would nod in sync with \emph{jump} actions or tilt their head to the direction of movement in response to avoiding an enemy. Clustering players with respect to expressivity and performance also identified interesting patterns: for instance, expert players were either very expressive, in the sense that they were immersed in the game action and mimicked their character's movement with body movements, or almost inanimate, indicating a high level of concentration. Overall, this work identified the need to combine player behaviour with expressive analysis \cite{kotsia2016multimodal} so as to produce meaningful and dependable results regarding player engagement. In the same framework, Pedersen et al.  \cite{pedersen2010modeling} produced levels for IMB which were predicted to be fun and engaging for each particular player, based on their affective and behavioural input while playing. This concept combines the sensing and player experience prediction work with modelling the aspects of the game level that make it \emph{hard}, \emph{fun} or \emph{irritating} for each player: in the context of platform games, such as SMB or IMB, these factors include the number of gaps in a given level, the gap size, the number and placement of enemies and the positioning of rewards and power-ups. This work showed that by mapping player experience modelling with difficulty modelling, content generation algorithms can create individual game levels with a high degree of probability to be interesting and engaging.

Besides affective or audiovisual data of people playing games, the most relevant source of data comes from player behaviour (actions within the game world). The amount of data produced by players during gameplay differs with each genre, with turn-based or strategy games producing a few samples per minute and action games or Real-time Strategy (RTS) games providing hundreds of individual player actions per minute (APMs): in ``Starcraft'', one of the most popular RTS games ever produced, top players typically record around 400 APMs in the preparation phase and close to 800 APMs during battle. As a result, accumulating large corpora of data and using them to train different AI/ML architectures has been a very popular use case for games and ML researchers. For example, Ravari et al. \cite{ravari2016starcraft} utilise the datasets presented in \cite{robertson2014improved} and \cite{synnaeve2012dataset} to predict the winner of each match, identifying relevant and important time-dependent (e.g. player actions, such as \emph{build} or \emph{attack}) and time-independent features, such as buildable areas in each game map or height of specific areas; to achieve this, they employ Gradient Boosting Regression Trees (GBRT) \cite{friedman2002stochastic} and Random Forest (RF) \cite{breiman2001random} implementations in Scikit-learn, an open-source Python package. In the same context, Lin et al. produced a very large dataset of more than 65.000 games, which also includes visual information, besides player behaviour \cite{lin2017stardata}; the sheer amount of data included here (1535 million frames, 496 million player actions) illustrates the relevance of player behaviour data to Big Data algorithms and processes (cf. \cite{bertens2017games} on churn prediction, i.e. when players quit playing a particular game, or \cite{willson2015zynga} on how player behaviour data are processed in the game industry to promote player experience and spending behaviour)  

A more contemporary source of data to be used with AI/ML algorithms comes from players interacting with other players during or before gameplay, in the form of text conversation (chat) or using voice. As mentioned earlier, gameplay eliminates most of the social inhibitions in players and allows for richer interaction and a wider variety of extreme emotions. Murnion et al. \cite{murnion2018machine} utilise commercial sentiment analysis tools, such as Twinword Sentiment Analysis and Microsoft Azure Cognitive Services, to process player behaviour and game logs from an online multi-player game called World of Tanks (WoT); the authors are looking for positive vs. negative interactions and specific abusive behaviours, such as derogatory insults or racist attacks. An important aspect of this work is that it uses easily accessible services to retrieve, decrypt and process the data, making the study easy to replicate and extend. The context of the study is also very interesting, since cyberbullying can result to extremely negative emotions and decisions, both with respect to players' real lives, as well as their sense of attrition from the game: different studies have shown that more than 50\% of players have either quit or considered quitting a game because of cyberbullying behaviours \cite{fryling2015cyberbullying}. A similar approach regarding game data was used in \cite{musabirov2018between}, where authors used topic modelling and statistical analysis to analyse interactions between viewers (or spectators) of matches in Dota\footnote{A richer dataset that includes 50000 matches, game data, player skill ratings and chat can be found at https://www.kaggle.com/devinanzelmo/dota-2-matches}, a real-time multiplayer battle game. Their work identified patterns similar to those of football match spectators, especially in the case of intra-audience effects, despite the fact that audience commentary in eSports does not reach (and, hence, influence) the players directly. 

\section{Intelligent game content generation and selection}
Game content generation has been a very active field in which AI/ML showcase their potential when it comes to generating data, mainly since the term \emph{content} can refer to most audio, visual and narrative concepts in a game. Besides visual appearance, such as environment aesthetics or 2D/3D models of characters and environments, game content may refer to audio or aural media (e.g. the soundtrack of the game or specific audio effects used in response to game events, such as firing a weapon or player death), graphical user interfaces (GUI) \cite{popp2013tool}, where interactive elements are used by the game to convey information (e.g. that another player or enemy is nearby) or by players to select game options and engage in game behaviour, or even the game narrative itself. The latter case (\cite{imabuchi2012story}, \cite{ogata2015building}, \cite{robertson2015automated}) is extremely interesting, since it caters for the generation of different stories and games, based on an initial plan or narrative by the game designer. The success of narrative generation techniques in Role-Playing Games (RPGs), which are mainly popular in Japan, shows that this content generation option has the potential to create longer-lasting gameplay and keep players motivated, serving the needs of both researchers and the industry \cite{smith2011pcg}. A special case of narrative generation includes NPC planning and behaviour \cite{savidis2018there}, with content generation algorithms choosing and executing non-player character actions and dialogues based on virtual personalities (e.g. a \emph{sidekick} or an enemy) or player behaviour (e.g. respond to the player ransacking a hut which belongs to a friendly villager in an RPG).

\begin{figure}[h!]
    \centering
    \includegraphics[width=0.8\textwidth]{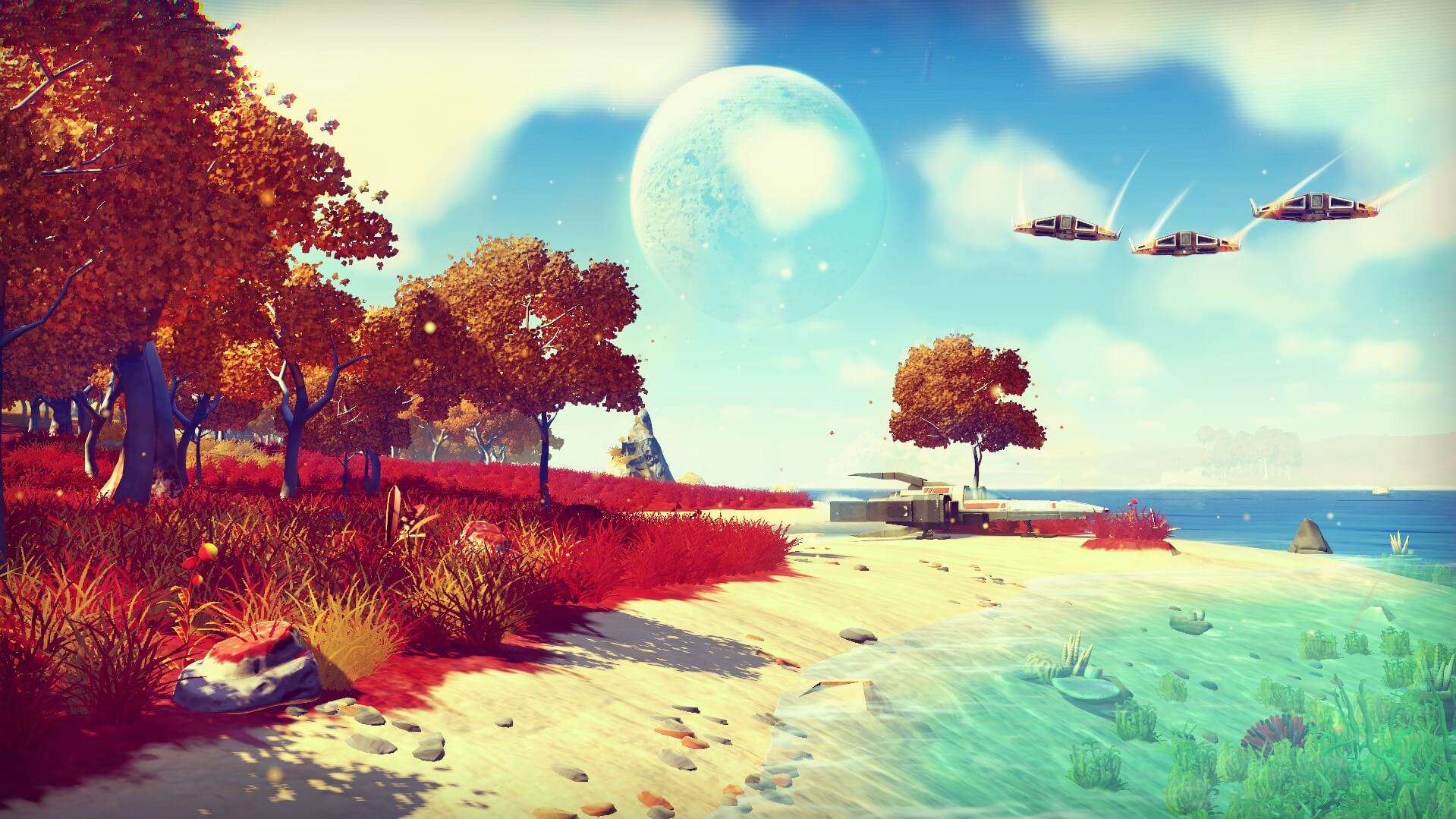}
    \caption{Planet and environment generation in No Man's Sky}
    \label{NMS}
\end{figure}

Intelligent content generation techniques can also be categorised with respect to the level of automation they require or provide. Some of those techniques enable designers to be involved in the process, either by initiating major changes in the way generated content is evolved \cite{liapis2016mixed} or by allowing players to adapt the content generated in the game \cite{shaker2012towards}. Fully automated techniques \cite{shaker2016procedural}, usually referred to as Procedural Content Generation (PCG), have been extremely popular with researchers, since they require little or no input, besides fine-tuning the respective algorithm parameters, but they also enjoy success in commercial games: in the 1980's, dungeon games such as \emph{Akalabeth} and \emph{Rogue} were among the first to use automatic (but sometimes random) content generation, while \emph{Elite} (1985), a 3D space exploration game, used content generation to author 8 galaxies with 256 solar systems each and 1 to 12 planets in each solar system, all within 32Kb of code. Between \emph{Diablo} (1995) and the recent years, PCG was mostly constrained to RPGs and dungeon layouts, with automatic content creation being revived by Minecraft (2011), developed by Mojang and now owned by Microsoft. More recent games include \emph{Left for Dead} (2008) (instantiating game objects such as trees, monsters or treasures), \emph{S.T.A.L.K.E.R.: The Shadow of Chernobyl} (2007) (dynamic systems create unscripted NPC behaviour), \emph{Apophenia} (2008) (generation of puzzles and plots), and mainly \emph{No Man's Sky} (2016) with a procedurally generated deterministic open world universe and planets with unique flora and fauna (Figure \ref{NMS}), and various sentient alien species.

\begin{figure}[h!]
    \centering
    \includegraphics[width=0.8\textwidth]{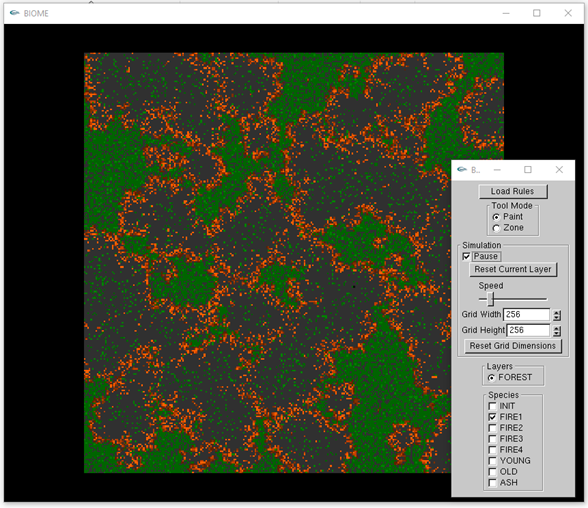}
    \caption{Generated forest fire in BIOME}
    \label{biome}
\end{figure}

Automatic PCG is usually matched with a respective generation algorithm and relevant constraints that match the game design or simply make sense. Vocabulary- or grammar-based generation algorithms are usually deployed to create game worlds, mazes or dungeons (\cite{johnson2010cellular}) and plants, whose recursive shape lends well to the way generation works in cellular automata \cite{miremadi2011bdd} or L-systems \cite{togelius2016grammars}. More recently, M{\"u}ller et al. used L-systems to \cite{muller2006procedural} to create 3D buildings with different sizes, number of rooms or floors. This is an interesting example, since the way the L-system is initially authored may reflect specific rules or constraints (e.g. to create a variety of structurally sound buildings in \cite{whiting2009procedural}) or even be used to generate complete cities \cite{talton2011metropolis}). Given the amount of polygonal geometry and texture images each building may require in a 3D environment, an automatic PCG algorithm that is able to create huge worlds with very little overhead in memory or CPU usage can replace the need to author the necessary objects beforehand. This is also the case with algorithms based on Artifical Life approaches: BIOME\footnote{Download BIOME from http://www.spore.com/comm/prototypes} is a programmable cellular automata simulator that allows users to develop simple ``SimCity-like'' grids, simulating phenomena such as forest fires, disease epidemics or animal migration patterns (cf. Figure \ref{biome}).

When it comes to \emph{sequences} of data to be generated, for instance in procedural music generation or when planning behaviours for NPCs, Hidden Markov Models (HMMs) are a usual choice. Snodgrass et al. \cite{snodgrass2016learning} generated sequences of game levels for different platform games, comparing their performance in each one, while Plans et al. \cite{plans2012experience} combined generation with player experience to author the music score for a game. More recently, Long Short-Term Memory (LSTM) networks have been used to generate game levels, e.g. in a Super Mario Bros. clone \cite{summerville2018procedural}. LSTMs seem to have taken over the PCGML (Procedural Content Generation based on Machine Learning) experiments, thanks to the readily available implementations in Python and C\#, as well as their recurrent nature that caters for generation of diverse and (theoretically) infinite content \cite{risi2019procedural}. For instance, Savery and Weinberg \cite{savery2020shimon} used LSTMs to synthesise musical scores based on image and video analysis, and Botoni et al. \cite{bottoni2020character} to create NPCs with more depth in terms of dialogue and style.

\subsection{Generating content for a language education game}
Generation of appropriate content for serious/educational games is an extremely important concept since it can make all the difference between adoption and retainment of the game, thus increasing the possibility to achieve its learning objectives, and attrition \cite{virvou2005combining}. In the iRead project\footnote{iRead project, https://iread-project.eu/}, we are creating a serious game and supporting applications for entry-level language learning of English, English as a Foreign Language (EFL), German, Spanish and Greek \cite{panagiotopoylos2020iread}. The core software applications developed in the project are a reader application\footnote{Amigo reader application, https://iread-project.eu/amigo-reader/}, which highlights parts of the words contained in the text, given specific criteria, and a serious game\footnote{Navigo game, https://iread-project.eu/game/}, which consists of a series of gamified activities utilising words and sentences. The foundation of these applications and the software infrastructure that provides access to the content consists of language models for each language, including for children with dyslexia; following the definition of extensive phonological and syntactic models for these languages, the linguists in the project worked with teachers to define the learning objectives for each of the target age groups, as well as the sequence in which each language feature should be taught \cite{mavrikis2019towards}. The sequencing of these features, including which prerequisites should be taught and mastered by the students before moving on to more advanced features, was encoded in a tree-like hierarchical graph; essentially, this graph encapsulates both the language model (i.e. the features that make up each word or sentence, at least at the given language level) and the teaching model, represented by the selection of necessary features for each school year and the succession in which they should be taught. When a new student registers with the iRead system, this graph is instantiated as a user profile, with different values of mastery for each feature, depending on the student's age.

\begin{figure}[h]
    \centering
    \includegraphics[width=0.9\textwidth]{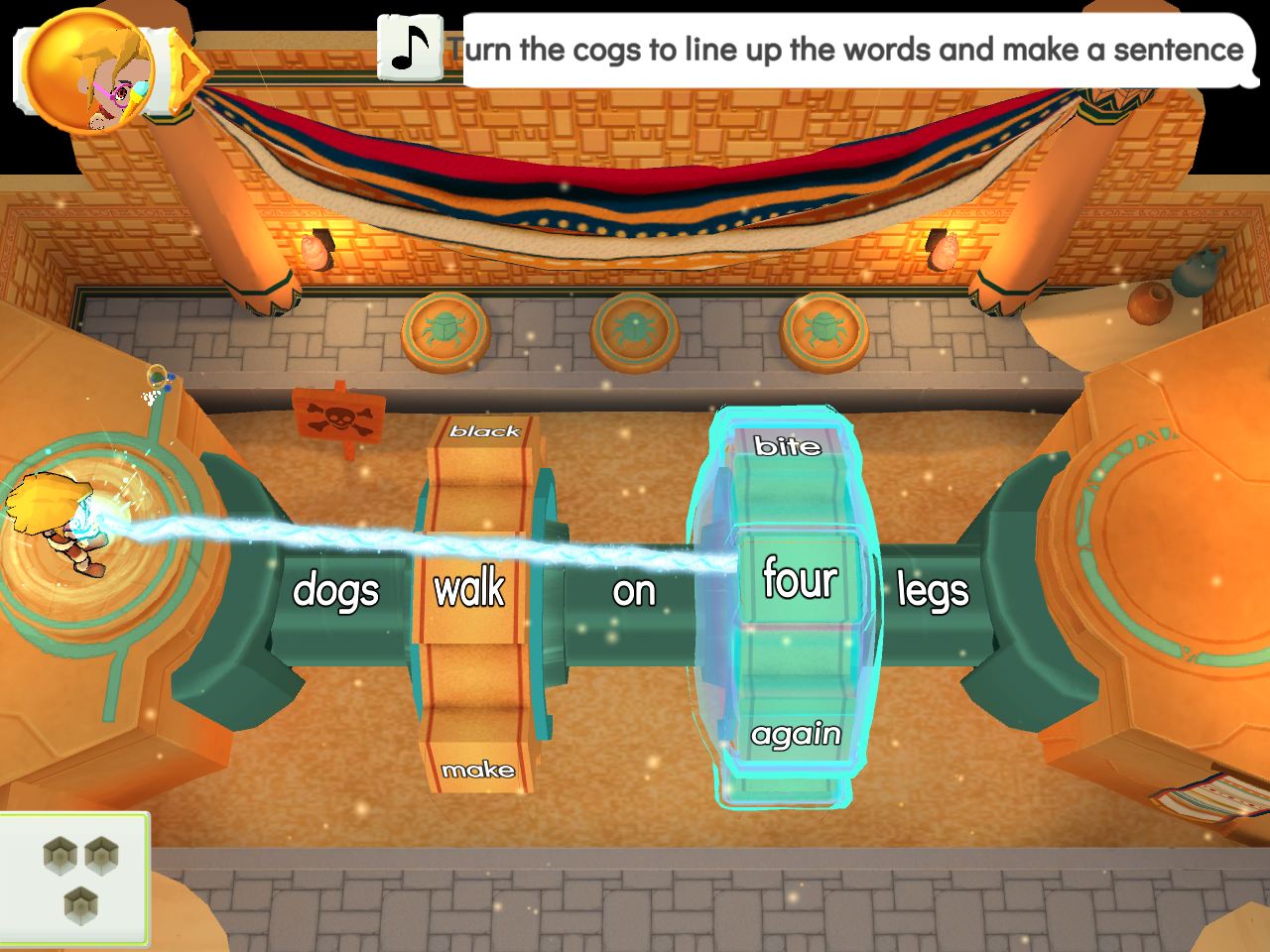}
    \caption{Game content generation for a Navigo mini-game}
    \label{sentences}
\end{figure}

This is where the adaptive content generation component \cite{tsatiris2020developing} in iRead kicks in, first by utilising the mastery levels for each feature to select proper content from the project resource engine (dictionaries and texts) and then by updating the student’s model based on their performance in each language game they play; when the mastery level for a given feature surpasses a selected threshold (75\%), subsequent features in the model hierarchy become available to play with, provided that all prerequisites for them have been met. In the context of iRead, the game content consists of selecting a particular game activity; a language feature to work with; and a set of words or a sentence that corresponds to that feature (e.g. a particular letter, phoneme or a sequence of phonemes).

\begin{figure}[h]
    \centering
    \includegraphics[width=0.75\textwidth]{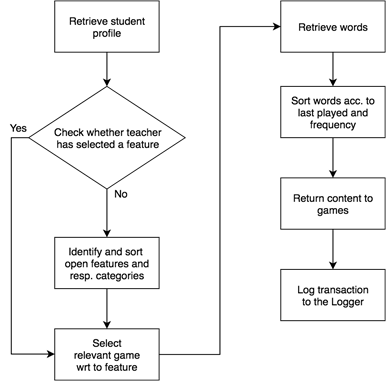}
    \caption{Flowchart of actions to generate content in iRead}
    \label{flowchart}
\end{figure}

The content selection process starts (cf. Figure \ref{flowchart} for an outline of the process) with the given student model, i.e. the mastery level for each open feature; then it selects the content for each session by filtering the available resources with a set of rules defined by project researchers after productive consultation sessions with the teachers collaborating with them. These rules were first defined in verbal form, in order to promote the teaching objectives of the games, with each of them corresponding to a particular pedagogical rationale. For instance, when multiple features are open (available to play with), the Adaptation component sorts them by taking into account how many times each of the features has been used in earlier games and how well the student has previously performed when presented with that feature. The reasoning here is that students should start from an easier feature and should not be playing a feature they have not done well with recently, thus fostering motivation and efficacy. Other rules attempt to reinforce learning by combining the feature mastery level achieved in previous games with the number of gameplay sessions since that feature was last used, and by reopening a fully mastered feature after ten games have been played since it was last used. The assumption is that the student has fully mastered that feature, but they must repeat it once in a while, to showcase their progress and long-term mastery. Finally, a number of feature selection rules deal with students not progressing as expected or not having truly mastered the language features which correspond to their age level: if a feature has been practised twice and the feature mastery is not improving, the mastery level for that feature and its prerequisites is reduced, so that both can be revisited in future sessions. This content selection strategy treats the assumption that students in a given age group have already mastered certain language features, by allowing them to go back to required knowledge, if there is no system evidence that it has been acquired. In addition to selecting proper word content, the iRead adaptivity system utilises a rule-based strategy to select specific game activities to utilise those words: if a feature has not been previously used in a game for the particular student, then the selected game should promote accuracy in using that language characteristic, before moving on to games which stimulate automaticity.

The second part of the adaptation component in iRead has to do with re-evaluating the value of mastery for the language feature used in a game. During the consultation sessions, teachers mentioned a number of requirements for this process: changes in mastery values should not be abrupt, especially when students make an occasional error in one of the activities; besides this, they should help students demonstrate complete mastery of a feature within a handful of gaming sessions, allowing them to move on to more advanced and interesting features. A mathematical definition that accommodates these requirements, while leaving room for experimentation and adjustments of the process, is that of Exponential Moving Average (EMA) \cite{pelanek2017experimental}: essentially, this takes into account previous attempts at a particular feature (previous game sessions) but gives more weight to recent attempts. The number of previous attempts to consider may be defined by each implementation; in iRead, we implemented the complete definition, but chose to consider only the previous value of mastery, when calculating the next one. This averaging process allows students to show complete mastery in just three games, since each newly opened feature is initialised with a value of 5 and after three perfect games reaches the maximum value of 10. In addition, in case the student makes one or more errors during game play, the respective mastery value may be reduced by 1 at maximum. This mechanic, along with the rules which prioritise features given the recent gameplay attempts, allows students to practise different language content, without being stuck with difficult features.

On-going evaluation (\cite{BUNTING2021100236}, \cite{Rvsz2020TheEO}) has shown that the automated content selection and the profile re-evaluation processes are quite close to what teachers expect and provide suitable and interesting content for the students. Even though the re-evaluation mechanic allows unlocking subsequent features quite easily, there have been reports that students are being given the same features to play with during numerous successive game sessions. However, after going through the system logs of gameplay results and mastery evaluations, we reached the conclusion that this reflects the design of the respective language model, which imposed a large number of prerequisites to be unlocked before moving on to more complex features. This effectively illustrates the interplay between the different iRead components: the features which describe each language model, the graph of prerequisites which describes the sequencing embedded in the learning process, the mastery levels for each feature which reflect student performance, and the adaptation and re-evaluation rules described above, which prioritise content to implement teaching objectives.

The large-scale evaluation phase in schools across Europe is already underway, with more than 2000 students taking part. Even though it has been disrupted by the pandemic and schools closing down, we expect gameplay logs to keep coming in from students playing the games at home. Processing these logs will allow us to revisit specific parts of the adaptation component, primarily the content selection rules and the mastery re-evaluation implementation.

\section{Conclusions}
The interplay between AI/ML algorithms and digital games has been in the forefront of scientific news and research outlets for the past few years. The main reason for this is that it fosters adaptive player experiences \cite{yannakakis2014guest}, which promote and even maximise fun and engagement. Besides this, AI/ML can be used to select and generate diverse and related game content, even from sources of Open and Big Data (cf. \cite{vargianniti2020using} for a Monopoly clone populated with Open Data \cite{theocharis2014ontology} to teach Big Data rankings and associations in the context of a primary school geography course or \cite{chiotaki2020open} for an approach that uses Open Data in a card game for environmental education), making games more relevant to everyday life. In this chapter, we outlined some of the more prominent approaches which combine AI/ML with game design concepts and player behaviour to provide information about player experience and generate content that's predicted to maximise engagement. We also described an approach to estimate player experience and engagement based on behaviour and affective expressivity during gameplay and an intelligent algorithm that generates language content for a serious game, based on player performance and learning/teaching objectives. As users provide more and richer input to AI/ML algorithms through explicit choices and gameplay, it is expected that this interplay will become even more meaningful and will be integrated in more applications, expanding into education, inclusion \cite{schmolz2017doing} and gamification (\cite{legaki2019using}, \cite{legaki2020effect}).

\subsubsection*{Acknowledgements}
This work has been partly funded by the iRead project which has received funding from the European Union’s Horizon 2020 Research and Innovation programme under Grant Agreement No. 731724.

\bibliographystyle{plain}  
\bibliography{biblio.bib}

\end{document}